%% file: vandemeent_arxiv_2013.tex
\documentclass[fleqn]{article}

\usepackage[]{graphicx} 
\usepackage{subfigure} 

\usepackage{natbib}

\usepackage{algorithm}
\usepackage{algorithmic}

\usepackage{amssymb,amsfonts,amsmath}

\usepackage{hyperref}

\usepackage{color}
\usepackage[usenames,dvipsnames]{xcolor}

\input{macros.tex}

\usepackage[accepted]{icml2013} 

\icmltitlerunning{Hierarchically-coupled hidden Markov models for learning kinetic rates from single-molecule data}

\begin{document} 

\twocolumn[
\icmltitle{Hierarchically-coupled hidden Markov models\\ for learning kinetic rates from single-molecule data}

\icmlauthor{Jan-Willem van de Meent$^1$}{janwillem.vandemeent@columbia.edu}
\icmlauthor{Jonathan E. Bronson$^1$}{jonathan.bronson@gmail.com}
\icmlauthor{Frank Wood$^2$}{fwood@robots.ox.ac.uk}
\icmlauthor{Ruben L. Gonzalez Jr.$^1$}{rlg2118@columbia.edu}
\icmlauthor{Chris H. Wiggins$^1$}{chris.wiggins@columbia.edu}
\icmladdress{$^1$Columbia University, New York, NY, USA; $^2$University of Oxford, Oxford, UK}

\icmlkeywords{hidden Markov models; variational Bayesian inference; empirical Bayes; type II maximum likelihood; single-molecule biophysics}

\vskip 0.3in
]

\begin{abstract} 
\input{sec_abstract.tex}
\end{abstract} 

\section{Introduction}
\input{sec_introduction.tex}

\section{Related Work}
\input{sec_related.tex}

\section{Variational empirical Bayes}
\input{sec_veb.tex}

\section{Performance on simulated data}
\input{sec_validation.tex}

\section{Analysis of smFRET data}
\input{sec_experiments.tex}

\section{Model Selection Criteria}
\input{sec_modelselection.tex}

\section{Discussion}
\input{sec_discussion.tex}

\section{Acknowledgements}

The authors would like to thank Matt Hoffman and David Blei for helpful discussions. This work was supported by an NSF CAREER Award (MCB 0644262) and an NIH-NIGMS grant (R01 GM084288) to R.L.G.; and a Rubicon fellowship from the Netherlands Organization for Scientific Research (NWO) to J.W.M.

\newpage
\bibliography{vandemeent_arxiv_2013}
\bibliographystyle{icml2013}

\end{document}

%% file: macros.tex
\newcommand{\argmax}{\operatornamewithlimits{argmax}}
\newcommand{\argmin}{\operatornamewithlimits{argmin}}
\newcommand{\ifrac}[2]{#1 \:/\: #2}
\newcommand{\Dkl}{D_{\rm KL}}

\newcommand{\Lvb}{L^{\rm vb}}
\newcommand{\Lveb}{L^{\rm veb}}

\newcommand{\jwm}[1]{#1}
\newcommand{\rlg}[1]{#1}
\newcommand{\fw}[1]{#1}
\newcommand{\wei}[1]{#1}

%% file: sec_abstract.tex

We address the problem of analyzing sets of noisy time-varying signals that all report on the same process but confound straightforward analyses due to complex inter-signal heterogeneities and measurement artifacts.  In particular we consider single-molecule experiments which indirectly measure the distinct steps in a biomolecular process via observations of noisy time-dependent signals such as a fluorescence intensity or bead position. Straightforward hidden Markov model (HMM) analyses attempt to characterize such processes in terms of a set of conformational states, the transitions that can occur between these states, and the associated rates at which those transitions occur; but require ad-hoc post-processing steps to combine multiple signals.  Here we develop a hierarchically coupled HMM that allows experimentalists to deal with inter-signal variability in a principled and automatic way. Our approach is a generalized expectation maximization hyperparameter point estimation procedure with variational Bayes at the level of individual time series that learns an single interpretable representation of the overall data generating process.

%% file: sec_introduction.tex

\begin{figure*}
    \includegraphics[width=\textwidth]{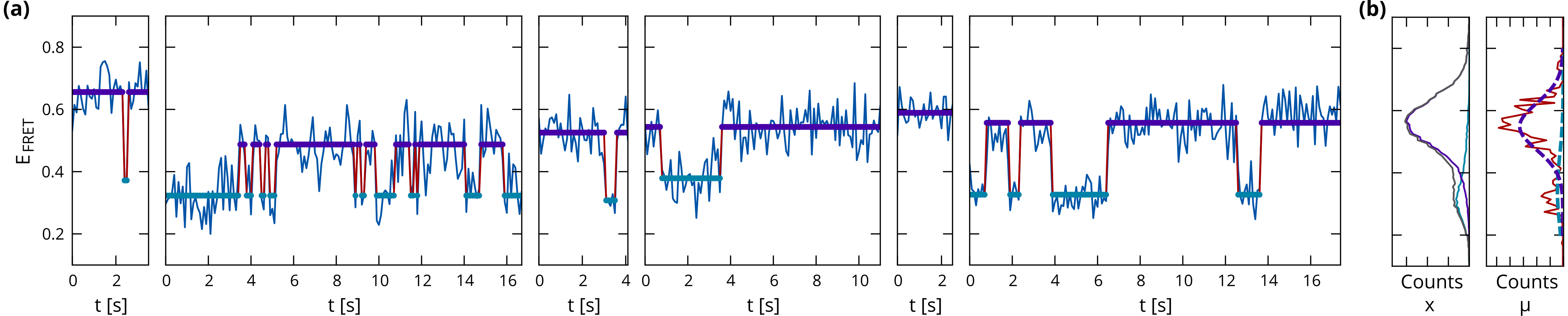}
    \caption{\label{fig_ex_traces} {\bf (a)} Sample time series from a single-molecule fluorescent energy transfer (smFRET) experiment, \jwm{which measures a ratio of intensities of two fluorescent labels $E_{\rm FRET} = I_{\rm A} / (I_{\rm D} + I_{\rm A})$ to detect conformational transitions between two states (see section \ref{sec:experiments}). Analysis with a variational Bayesian HMM shows states with means and precisions that vary significantly from molecule to molecule.} Moreover photobleaching of the fluorophores results in time series of variable (exponentially distributed) lengths. {\bf (b)} \jwm{Histogram of observables $x_{n,t}$, and inferred state means $\mu_{n,k}$ (red) along with empirical Bayes estimates of priors (dotted lines) for an ensemble of $N=336$ time series from a single experiment.}}
\end{figure*}

Over the past two decades, single-molecule biophysical techniques have revolutionized the study of Nature's biomolecular machines by enabling direct observation of some of the cell's most fundamental and complex biochemical reactions \cite{tinoco_gd_2012,joo_arbc_2008,borgia_arbc_2008,neuman_nm_2008,cornish_acscb_2007}. 
At the molecular level, such reactions can be described as a `kinetic scheme' that details the number of conformational states accessible to the biomolecular machine, the order in which these states are explored, and the associated transition rates.
In single-molecule biophysics, the goal of inference is to reconstruct this kinetic scheme from a noisy time-dependent observable, such as a fluorescence intensity or bead position, which indirectly reports on the state transitions in a biomolecular process.

The learning task that motivates this work is the estimation of a consensus kinetic scheme from a multitude of time series.
Single-molecule experiments commonly yield observations for an `ensemble' of hundreds of molecules.
An illustration of this analysis problem can be seen \jwm{in} figure \ref{fig_ex_traces}.
The time series shown are obtained in a single-molecule \rlg{fluorescence resonance energy transfer (smFRET) experiment, where a fluctuating intensity ratio $E_{\rm FRET} = I_A / (I_D + I_A)$ of two fluorescent molecules, known as the donor and acceptor, reports on the conformational transitions of a molecule of interest.}
Conditioned on the conformational state these observables are approximately normally distributed, but their means vary significantly over the ensemble owing to a combination of image-processing artifacts and physical inhomogeneities.
Moreover, the number of conformational transitions that can be observed for each molecule is limited by the life-time of the fluorescent labels, which have a fixed probability of photobleaching upon each excitation.
Learning a common set of states and associated transition probabilities from time series such as in figure \ref{fig_ex_traces} remains a challenging task.
Existing approaches use maximum likelihood \cite{mckinney_bpj_2006,greenfeld_plos1_2012} or variational Bayesian \cite{bronson_bpj_2009} estimation on HMMs to infer independent models for each individual time series, resulting in a set of similar but variable parameter estimates.
Experimentalists then resort to an ad-hoc semi-manual binning of states with similar observation means, after which the corresponding binned transition counts can be averaged over the ensemble to obtain a more informed estimate of the consensus kinetic rates, implicitly assuming an identical set of transition probabilities for all time series.

The contribution presented here is to develop hierarchically coupled HMMs to learn a distribution on the parameters for each state. 
Specifically, we assume a HMM joint $p(x_n, z_n \,|\, \theta_n)$ for the observables $x_n$ and latent states $z_n$ for each time series $n \in [N]$, conditioned on a set of parameters $\theta_n$ drawn from a shared prior $p(\theta \,|\, \psi)$.
In the statistical community this is known as a conditionally independent hierarchical model \cite{kass_jasa_1989}.
The hyperparameters are estimated with an empirical Bayes approach \cite{morris_jasa_1983}.
This generalized expectation (EM) procedure iteratively performs variational Bayes (VB) estimation for each time series, after which the summed lower bound is maximized with respect to $\psi$.

The resulting procedure, which we will call variational empirical Bayes (VEB), infers a single consensus parameter distribution from an ensemble of time series that represents the kinetic scheme that is of experimental interest.
The VEB procedure also yields more accurate inference results in individual time series.
This is a well-known feature of empirical Bayes models \cite{berger_jasa_1982}, that intuitively follows from the fact that the shared prior $p(\theta \,|\, \psi)$ incorporates knowledge of parameter values across the ensemble that can aid the inference process in individual time series.
Finally our hierarchical construction allows comparison of prior densities to their ensemble-averaged posterior estimates, providing the experimentalist with a  intuitive diagnostic for the agreement between observed experimental data and a chosen graphical model, which can then inform the next iteration of model design.

%% file: sec_related.tex

Variational approaches are a mainstay of Bayesian inference \cite{jordan_ml_1999,wainwright_ftml_2008,bishop_book_2006}.
Variational Bayes estimation for HMMs has previously been applied to smFRET experiments \cite{bronson_bpj_2009,fei_pnas_2009,bronson_bmcbi_2010}, and these techniques have both been used directly \cite{sorgenfrei_nn_2011} and in a modified form \cite{okamoto_bpj_2012} in the context of several other single-molecule experiments.

Methods that obtain point estimates for a set of hyperparameters are known under a variety of names, including empirical Bayes, type II maximum likelihood, generalized maximum likelihood and evidence approximation \cite{bishop_book_2006}.
A common use for such techniques is to adaptively set the hyperparameters of a model to values that are appropriate for a given inference application.
These approaches have the well-documented feature of allowing more accurate inferences for individual samples in learning scenarios where parameters are correlated \cite{morris_jasa_1983,kass_jasa_1989,carlin_book_1996}, which can be interpreted as a generalization of the Stein effect \cite{casella_as_1985,stein_as_1981,berger_jasa_1982}.
In some cases, such as mixture models, type II maximum likelihood methods can also be employed in quasi non-parametric manner, by using a larger than required number of mixture components and relying on the inference procedure to leave superfluous components unpopulated \cite{corduneanu_aistats_2001}.

\begin{figure}[!t]
    \includegraphics[width=\columnwidth]{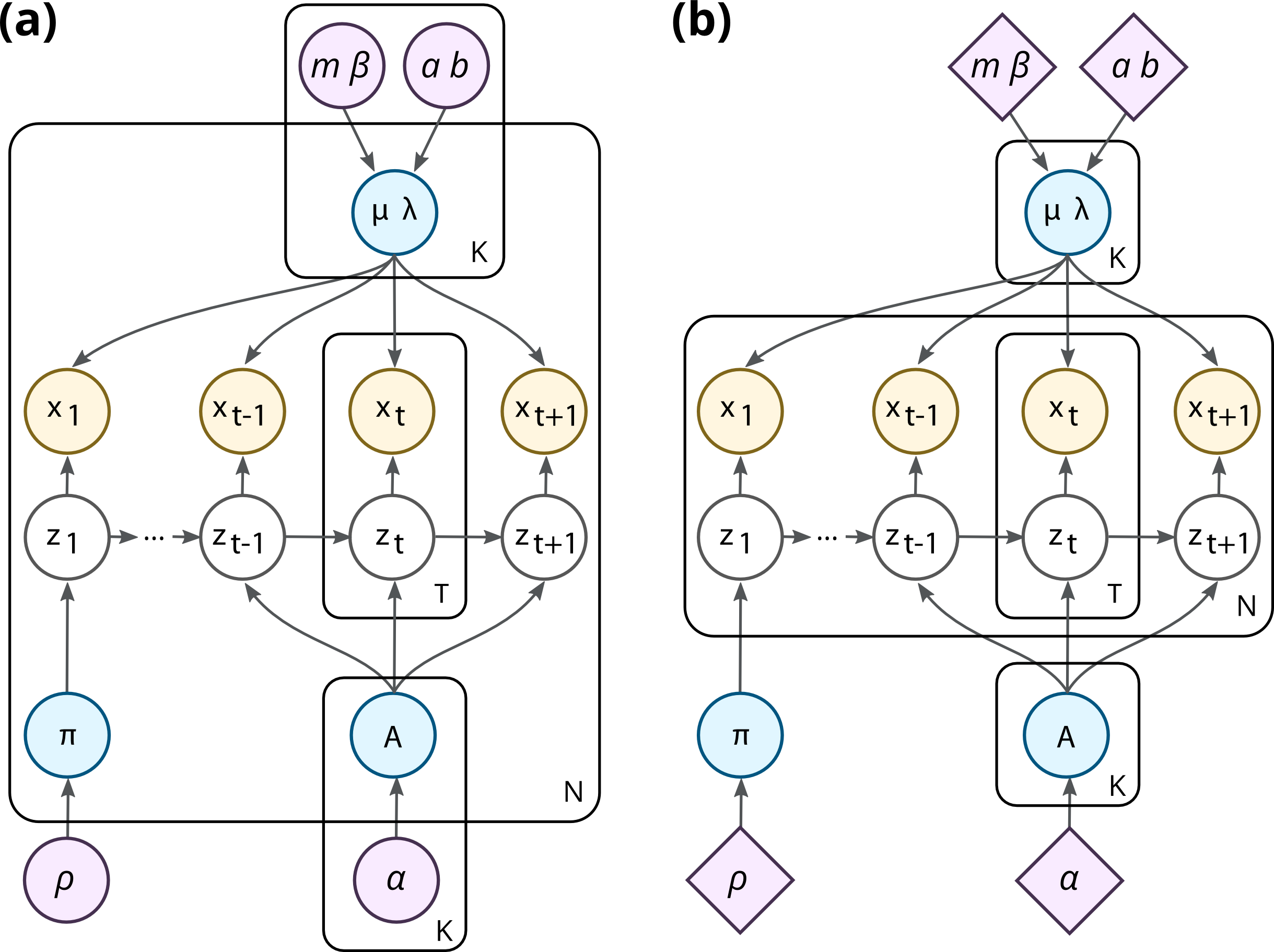}
    \caption{\label{fig_gm_comp}
        HMMs for ensembles of time series. (a) Our model employs a separate set of hyperparameters for each state, representing a consensus model whose parameters are allowed to vary across the ensemble. (b) A fully shared approach with a single set of $K$ states, which are identical for all $N$ time series, and are each drawn from the same (typically uninformative) prior.
    }
\end{figure}

In this biophysical application, the hyperparameters provide an explicit representation of the common features in multiple time series, which are individually represented as HMMs with Gaussian observations
\begin{align*}
    x_{n,t} \,|\, z_{n,t}\!=\!k
    &~\sim~ 
    {\rm Normal}(\mu_{n,k}, \lambda_{n,k}) 
    ,
    \\
    z_{n,t} \,|\, z_{n,t-1}\!=\!k
    &~\sim~ 
    {\rm Discrete}(A_{n,k})
    ,
    \\
    z_{n,0}
    &~\sim~ 
    {\rm Discrete}(\pi_n)
    .
\end{align*}
For each $n \in [N]$ time series, the set of parameters $\theta_n = \{\mu_n, \lambda_n, A_n, \pi_n\}$ are drawn from the priors
\begin{align*}
    \mu_{n,k} 
    &~\sim~ 
    {\rm Normal}(m_k, \beta_k\lambda_{n,k}) 
    ,
    \\
    \lambda_{n,k} 
    &~\sim~ 
    {\rm Gamma}(a_k, b_k) 
    ,
    \\
    A_{n,k} 
    &~\sim~ 
    {\rm Dirichlet}(\alpha_k) 
    ,
    \\
    \pi_{n} 
    &~\sim~ 
    {\rm Dirichlet}(\rho) 
    .
\end{align*}
This graphical model (fig \ref{fig_gm_comp}a) represents each of the $k \in [K]$ consensus states with a set hyperparameters $\psi = \{m_k, \beta_k, a_k, b_k, \alpha_k, \rho\}$, introducing a hierarchical coupling that can be used to represent an ensemble of HMMs with similar yet not identical parameters.
This type of construction is more expressive than a model with identical parameters for each time series (fig \ref{fig_gm_comp}b).
At the same time the use of differentiated priors defines a correspondence between the states in individual time series and a set of consensus states, eliminating label-switching issues across the ensemble.

As we will see in our discussion of results, estimation of the hyperparameters allows us to take advantage of the known features of empirical Bayes methods, including increased accuracy of inferred parameters and adaptive selection of model complexity.
However, from the point of view of modeling single-molecule biophysical systems the most substantial advantage of this type of approach is one of dimensionality reduction. 
A comparatively small set of $K(K+5)$ hyperparameters provides a single model that incorporates the information contained in hundreds of single-molecule time series in a statistically robust manner.
In contrast to applications where empirical Bayes estimation is used to aid posterior inference, the learned hyperparameters in this hierarchical model are in fact the primary quantity of interest.

%% file: sec_veb.tex

All variants of expectation maximization procedures have equivalent formulations in terms of the minimization of a Kullback-Leibler (KL) divergence, or the maximization of a (lower bound marginal) likelihood. 
In the first formulation an optimization criterion for parametric empirical Bayes models can be written as
\begin{equation}
    \label{eq_eb_criterion}
    \psi^*
    =
    \argmin_{\psi}
    \Dkl 
    \big[
        p(z, \theta \,|\, x, \psi) 
        \:||\: 
        p(z, \theta \,|\, \psi)
    \big] 
    ,
\end{equation}
Intuitively this self-consistency relationship states that the joint prior $p(z,\theta \,|\, \psi^*)$  should \fw{be} chosen to be as similar as possible to the posterior $p(z, \theta \,|\, x, \psi^*)$.

A dual problem for equation \ref{eq_eb_criterion} can be constructed by introduction of a variational approximation to the posterior.
Given any function $q(z,\theta)$, the marginal likelihood $p(x\,|\,\psi)$, also known as the evidence, can be trivially represented as
\begin{align*}
    p(x \,|\, \psi)
    &=
    \frac{p(x,z,\theta \,|\, \psi)}{q(z,\theta)}
    \frac{q(z,\theta)}{p(z,\theta \:|\; x, \psi)} 
    .
\end{align*}
Taking the logarithm of the above equation, followed by an expectation over
$q(z,\theta)$ yields the relationship
\begin{align*}
    \Lveb
    &=
    E_q
    \left[
        \log \frac{p(x, z, \theta \,|\, \psi)}{q(z,\theta)}
    \right]
    \\
    &=
    \log p(x \,|\, \psi)
    -
    \Dkl
    \big[
        q(z,\theta)
        \:||\:
        p(z,\theta \,|\, x, \psi)
    \big]
    .
\end{align*}
In a model where $p(x,z,\theta \,|\, \psi) = p(x \,|\, z, \theta) p(z, \theta \,|\, \psi)$ we can additionally write
\begin{align*}
    \Lveb
    =~
    &
    E_q
    \left[
        \log p(x \,|\, z, \theta)
    \right]
    \\
    &
    -
    \Dkl
    \big[
        q(z,\theta)
        \:||\:
        p(z,\theta \,|\, \psi)
    \big]
    .
\end{align*}
Because $\Lveb \le \log p(x \,|\, \psi)$, this quantity is often called an evidence lower bound (ELBO).
Maximization of $\Lveb$ with respect to $q(z,\theta)$ is equivalent to minimization of the KL-divergence between $q(z,\theta)$ and the posterior, whereas maximization with respect to $\psi$ minimizes the KL-divergence between $q(z,\theta)$ and the prior. 
These two steps can be combined to construct an iterative optimization algorithm 
\begin{align*}
    \ifrac{\delta \Lveb}{\delta q(z, \theta)} &= 0
    ,
    &
    \ifrac{\partial \Lveb}{\partial \psi} &= 0
    .
\end{align*}
If the posterior could be calculated directly, we could solve the first equation by setting $q(z,\theta) = p(z,\theta \,|\, x, \psi)$. The second equation then gives us the solution the hyperparameter estimation problem posed in equation \ref{eq_eb_criterion}. In this scenario, parametric empirical Bayes estimation is equivalent to an expectation maximization (EM) algorithm for the marginal likelihood $p(x \,|\, \psi)$.

For models where the posterior cannot be calculated directly we can substitute any analytically convenient form for $q(z,\theta)$ and calculate an approximation
\begin{align*}
    \label{eq_kl_q_pzq}
    q^*(z, \theta)
    &=
    \argmax_{q(z, \theta)}
    \Lveb[q(z, \theta), \psi]
    \\
    &=
    \argmin_{q(z, \theta)}
    \Dkl
    \big[
        q(z,\theta)
        \:||\:
        p(z,\theta \,|\, x,\psi) 
    \big]
    .
\end{align*}
An often convenient choice is a factorized form
\[
    q(z, \theta) 
    = 
    \prod_{n=1}^N 
    q(z_n) q(\theta_n)
    ,
\]
which in conjugate exponential family models guarantees that the approximate posterior has the same analytical form as the prior, i.e.~$q(\theta_n) = p(\theta_n \,|\, \hat \psi_n)$, where $\hat \psi_n$ is a set of variational posterior parameters that can typically be expressed as a weighted average of the hyperparameters and the expectation of a set of sufficient statistics \cite{beal_thesis_2003}.

With the above factorization, we can construct an approximate inference procedure for parametric empirical Bayes models, which we call variational empirical Bayes (VEB), that optimizes $\Lveb$ through
\begin{align*}
    \frac{\delta \Lveb}{\delta q(z_n)} &= 0
    ,
    &
    \frac{\delta \Lveb}{\delta q(\theta_n)} &= 0
    ,
    &
    \frac{\partial \Lveb}{\partial \psi} &= 0
    .
\end{align*}

This method is directly related to existing variational inference techniques on HMMs, which employ the lower bound
\begin{align*}
    \Lvb_n
    =~&
    E_q 
    \left[
        \frac{p(x_n, z_n, \theta_n \,|\, \psi)}{q(z_n)q(\theta_n)}
    \right]
    \\
    =~&
    \log p(x_n \,|\, \psi)
    \\ 
    &-
    \Dkl
    \big[
        q(z_n)q(\theta_n)
        \:||\:
        p(z_n, \theta_n \,|\, x_n, \psi)
    \big]
    .
\end{align*}
Because $\Lveb = \sum_n \Lvb_n$ optimization w.r.t.~$q(z,\theta)$ simply reduces to performing VB estimation on each of the individual time series.

The hyperparameter updates $\ifrac{\partial \Lveb}{\partial \psi} = 0$ take the form of a coupled set of equations over the $n \in [N]$ time series in the ensemble
\[ 
    0 
    = 
    \sum_n 
    E_{q(\theta_n)}
    \left[
        \frac{\partial \log p(\theta_n \,|\, \psi)}{\partial \psi}
    \right].
\]
We can express these coupled equations in terms of a set of ensemble averages, defined by
\[
    \bar{E}_q [ \:\cdot\: ]
    = 
    1/N \sum_n
    E_{q(\theta_n)} [ \:\cdot\: ]
    .
\]
For a Dirichlet prior, the hyperparameter updates simply match the log expectation values
\begin{align*}
    E_p[ \log A_k]
    &=
    \bar{E}_q[ \log A_k]
    ,
    \\
    E_p[ \log \pi]
    &=
    \bar{E}_q[ \log \pi]
    .
\end{align*}
In terms of $\psi$ these log expectation values can be expressed in terms of the digamma function $\Psi$
\begin{align*}
    &
    \Psi
    \big[
        {\textstyle \sum_m} \alpha_{km}
    \big]
    -
    \Psi
    \big[
        \alpha_{kl}
    \big] 
    \\
    &
    =
    \frac{1}{N}
    \sum_n
    \Psi
    \big[
        {\textstyle \sum_m} \hat{\alpha}_{n,km}
    \big]
    -
    \Psi
    \big[
        \hat{\alpha}_{n,kl}
    \big]
    .
\end{align*}
While these equations have no analytical solution their stationary point can be found efficiently with a Newton iteration method \cite{minka_report_2000}.

Maximimzation with respect to the Normal-Gamma prior on $\mu$ and $\lambda$ similarly yields four relationships between prior expectation values and their ensemble posterior averages
\begin{align*}
    E_{p}[\mu_k]
    &=
    \ifrac{\bar{E}_{q} [\mu_{k} \lambda_{k}]}
            {\bar{E}_{q} [\lambda_{k}]}    
    ,
    \\
    E_{p}[\lambda_k (\mu_k - m_k)^2]
    &=
    \bar{E}_{q}[\lambda_k (\mu_k - m_k)^2] 
    ,
    \\
    E_{p}[\log \lambda_k]
    &=
    \bar{E}_{q}[\log \lambda_k] 
    ,
    \\
    E_{p}[\lambda_k]
    &=
    \bar{E}_{q}[\lambda_k] 
    .
\end{align*}
which are equivalent to the following updates for $\psi$
\begin{align*}
    m_k 
    &=
    \bar{E}_q[\mu_k \lambda_k] 
    / 
    \bar{E}_q[\lambda_k]
    ,
    \\
    1 / \beta_k 
    &= 
    \bar{E}_{q}[\mu_k^2 \lambda_k] 
     - \bar{E}_q[\lambda_k \mu_k]^2 
       / \bar{E}_q[\lambda_k]
    ,
    \\
    \Psi(a_k) 
    - 
    \log(a_k)
    &=
    \bar{E}[\log \lambda_k]
    - 
    \log \bar{E}_q[\lambda_k]
    ,
    \\
    b_k
    &=
    a_k 
    /
    \bar{E}_q[\lambda_k]
    ,
\end{align*}
where the ensemble expectations are given by
\begin{align*}
    \bar{E}_q [ \lambda_k ]
    &=
    1/N
    {\textstyle \sum_n}
    \hat{a}_{n,k} 
    / 
    \hat{b}_{n,k} 
    ,
    \\
    \bar{E}_q [ \log \lambda_k ]
    &=
    1/N
    {\textstyle \sum_n}
    \psi(\hat{a}_{n,k})
    -
    \log \hat{b}_{n,k} 
    ,
    \\
    \bar{E}_q [ \mu_k \lambda_k ]
    &=
    1/N
    {\textstyle \sum_n}
    \hat{m}_{n,k} 
    \hat{a}_{n,k} 
    / 
    \hat{b}_{n,k} 
    ,
    \\
    \bar{E}_q [ \mu_k^2 \lambda_k ]
    &=
    1/N
    {\textstyle \sum_n}
    1 / \hat{\beta}_{n,k} 
    +
    (\hat{m}_{n,k})^2
    \hat{a}_{n,k} 
    / 
    \hat{b}_{n,k} 
    .
\end{align*}

%% file: sec_validation.tex

\label{sec:validation}

\begin{figure*}
    \noindent\includegraphics[width=\textwidth]{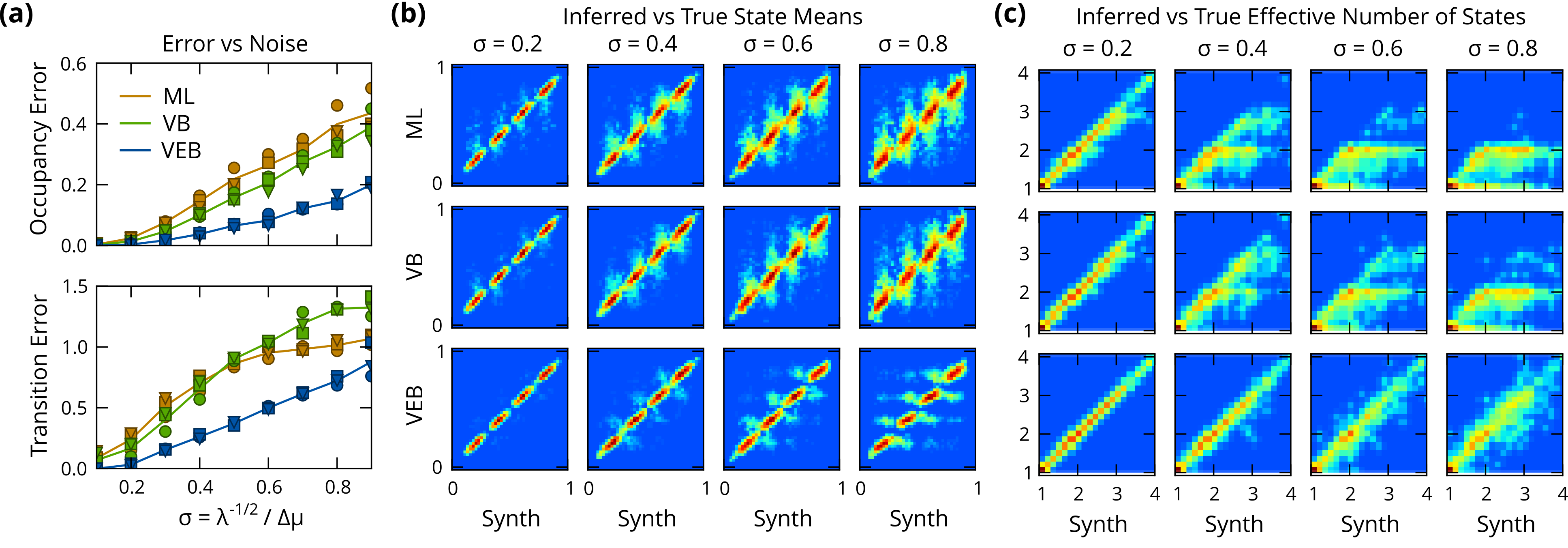}
    \caption{
    \label{fig_acc_vs_noise}
    Validation on simulated data. 
    {\bf (a)} \jwm{Errors in inferred transition pseudocounts, separated into diagonal `occupancy' and off-diagonal `transition' terms.}
    By both measures the VEB algorithm (blue) significantly outperforms VB (green) and ML (orange) methods. 
    {\bf (b)} Histograms (logarithmically scaled) of inferred vs true state mean $\mu_{z(t)}$, for simulated datasets with $K=4$. The increasing inference error is visible in the progressively blurred distribution along the vertical axis, which is markedly reduced when using VEB inference.
    {\bf (c)} Effective number of states. 
    ML and VB estimation (using the BIC and ELBO for model selection respectively), both systematically underestimate the number of states,
    whereas VEB algorithm shows a significantly improved performance, even at high noise.
    }
\end{figure*}

Because VEB estimation is a generalized EM algorithm for the hyperparameters, it is in principle subject to the usual concerns about overfitting and the number of restarts required to avoid local maxima.
At the same time results for type II maximum likelihood estimation of mixture weights suggest \cite{corduneanu_aistats_2001} that such estimation schemes may leave superfluous states unpopulated, and \jwm{the empirical Bayes literature suggests that} hyperparameter estimation results in more accurate posterior inference \jwm{\cite{morris_jasa_1983}}.
As a pragmatic approach to evaluating the performance of this method we have sampled simulated datasets from a range of experimentally plausible parameters, and compared inference results from the VEB algorithm to those obtained with maximum likelihood (ML) and variational Bayesian (VB) methods.

We performed inference on simulated datasets containing $N=500$ time series, each with $T=100$ time points, exploring $K = 3,4,5$ evenly spaced states at distance $\Delta \mu = 0.2$, and increase the noise relative to this separation. Noise levels for $\sigma = \lambda^{-1/2} / \Delta\mu$ range from $0.1$ (unrealistically noiseless) to $0.9$ (unrealistically noisy), where $0.5$ can be considered an upper limit of data that still permits analysis with state-of-the-art algorithms. 
In typical experiments, the variance of apparent state means $V_{p(\theta | \psi)}[\mu]$ is of a similar order of magnitude as the variance of the observables $V_{p(x | \theta,z)}[x]$.
For the purposes of this numerical experiment, we simulated a relatively homogeneous ensemble with $V_{p(\theta | \psi)}[\mu] = 0.4 \: V_{p(x | \theta,z)}[x]$ in order to isolate the influence of the experimental signal-to-noise ratio on the difficulty of the inference problem. 

In single-molecule biophysical applications, the goal of inference is to reconstruct a kinetic scheme in the form of a set of consensus states and associated transition probabilities. 
For this reason the quantities that are of primary interest are the transition pseudo-counts 
\[
    \xi_{n,ij} = \sum_t q(z_{n,t}\!=\!i, z_{n,t+1}\!=\!j) ~=~ \hat{\alpha}_{n,ij} - \alpha_{ij}
    .
\]
In ML and VB and estimation, the inferred states in each time series are unconnected, so we must construct some form of mapping $(n,i) \mapsto (n, k)$ to identify a set of consensus states. 
Here we use an approach that could typically be applied in existing experimental work flows, which is to simply cluster the inferred state means using a Gaussian mixture model. 
The remapped set of counts $\xi_{n,kl}$ can now be used to calculate the error relative to the true counts $\xi^0_{n,kl}$ that were sampled from the generative model.

We identify an `occupancy' error for the diagonal terms and a `transition' error for the off-diagonal terms, both defined in terms of the number of false positives and false negatives in the inferred transition counts.
These quantities have the advantage of having simple intuitive interpretations.
One is the error in the fraction of time spent in each state, whereas the other is the fraction of spurious/missed transitions.

The error rates obtained from this analysis are shown in figure \ref{fig_acc_vs_noise}a.
By both measures, the VEB algorithm significantly outperforms the ML and VB approaches over the entire range of noise levels. 
From an experimental perspective, this suggests that switching from a VB to a VEB procedure yields an improvement in accuracy comparable to increasing the signal-to-noise ratio of the measurements by a factor 2. 

Figure \ref{fig_acc_vs_noise}b shows 2-dimensional histograms (logarithmically scaled) that represent the joint distribution of inferred and true means $\mu_k$. 
Learning parameter distributions for each state results in well-defined peaks of the inferred posterior means, whereas the distributions for ML and VB become progressively poorly defined as the noise level increases. 

To evaluate the accuracy of inference in terms of the number of states populated in each time series, we calculate a quantity known as the effective number of degrees of freedom, which is based on the Shannon entropy of the time-averaged posterior 
\begin{align*}
    K^{\rm eff}
    &=
    \exp[H(Q(z))]
    &
    Q(z) 
    &= 
    \frac{1}{T} {\sum_t} q(z_t=z)
    .
\end{align*}
Note that for a discrete distribution with K choices, equal probability $Q(z) = 1/K$ for each outcome yields
\[
    \textstyle
    K^{\rm eff} 
    = \exp \left[ \sum_{k=1}^K  - 1/K \log(1/K) \right] 
    = K,
\]
whereas any distribution of lower entropy will have a correspondingly smaller number $K^{\rm eff}$, resulting in a continuous measure of the degrees of freedom as a function of state occupancy.

Figure \ref{fig_acc_vs_noise}c shows 2-dimensional histograms of the inferred and true number of effective states for the same dataset shown in figure \ref{fig_acc_vs_noise}b.
In VB and ML inference, we perform model selection on each time series using the lower bound $\Lvb$ and BIC respectively. 
VEB is given the correct maximum number of states at the hyperparameter level, but no additional model selection is performed for individual time series. 
The results in figure \ref{fig_acc_vs_noise}c show a clear picture: at high noise levels, both VB and ML estimation procedures underestimate the number of states, whereas VEB estimation is much more robust.

These results can each be seen as instances of the well known Stein effect \cite{berger_jasa_1982}: an estimator that exploits shared information between observations, in our case different time series, can provide more accurate inferences than an estimator that lacks this shared information.
Empirical Bayes methods obtain information about the distribution of the parameters $p(\theta \,|\, \psi)$ that is not  available in ML and VB approaches.
In other words, the prior knowledge encoded in the estimated hyperparameters can be used to obtain more robust inference results, as visible in the transition probabilities (fig \ref{fig_acc_vs_noise}a) and state means (fig \ref{fig_acc_vs_noise}b). 

We additionally observe that VEB inference populates the correct number of states in individual time series without needing model selection criteria (fig \ref{fig_acc_vs_noise}c), which is consistent with the results obtained for mixture models \cite{corduneanu_aistats_2001}.
A caveat is that all three analysis methods have thus far been supplied with a correct guess of the maximum number of states at the ensemble level.
We will return to the question of ensemble model selection in section \ref{sec:modelselection}.

%% file: sec_experiments.tex
\label{sec:experiments}
\begin{figure*}[!t]
    \noindent\includegraphics[width=\textwidth]{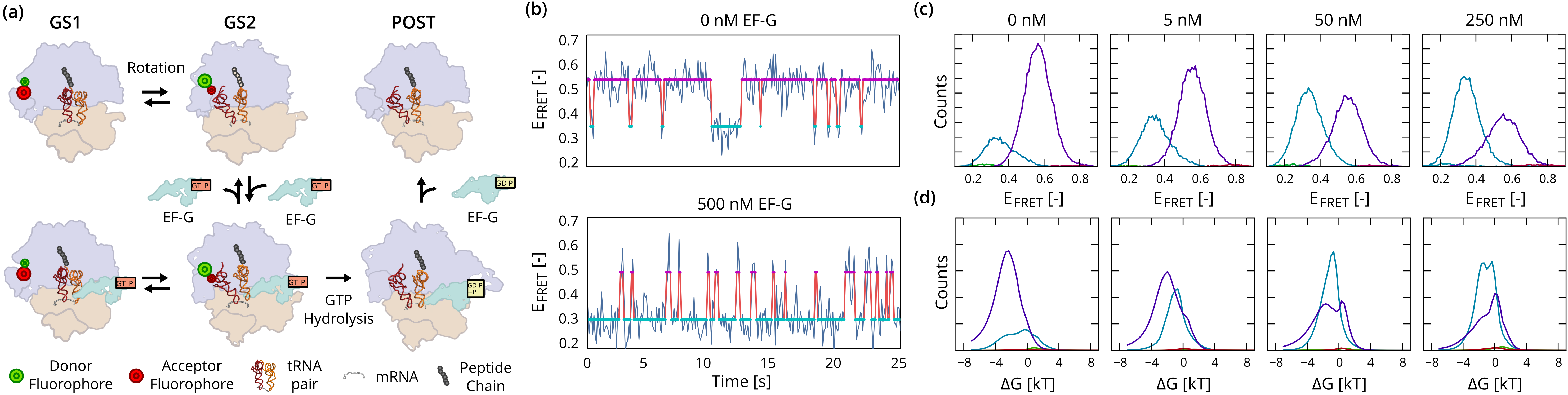}    
    \caption{smFRET studies of translocation in the bacterial ribosome \cite{fei_pnas_2009}. 
    {\label{fig_translocation_smfret}
    \bf (a)} The dominant pathway for translocation is believed to have three steps: A reversible rotation of the two subunits (purple and \rlg{tan}), followed by the binding of EF-G (green) which stabilizes the rotated GS2 state long enough for a GTP-driven transition to the post-translocation (POST) complex to take place. 
    {\bf (b)} smFRET signals show a shift of the equilibrium from the GS1 state \wei{(magenta, $E_{\rm FRET} \simeq 0.55$)} towards the GS2 state \wei{(cyan, $E_{\rm FRET} \simeq 0.35$)} in the presence of EF-G.
    {\bf (c)} Histograms of observables $x_{n,t}$, split by state, showing a continuous shift of the equilibrium towards GS2 as the concentration
    of EF-G is increased. 
    Out of 4 states used for inference, only two dominant states are significantly populated.
    {\bf (d)} Summed posterior distributions on the relative free energy $q(\Delta G_{n,k})$, showing a weak bi-modal distribution in the free energy of the GS2 state whose lower component becomes more populated with increasing EF-G concentration.
}
\end{figure*}

The VEB \jwm{method introduced} in this paper can be implemented for any single-molecule experiment amenable to analysis with HMMs and related graphical models. 
\jwm{Here} we focus on the analysis of smFRET experiments, which track conformational changes \rlg{in real time} by measuring the \rlg{anti-correlated intensity changes} of a pair of fluorophores, known as the donor and acceptor. \rlg{Figure \ref{fig_translocation_smfret} shows analysis of a set of experiments that investigate the mechanism of protein synthesis, or translation, by} the bacterial ribosome. 
\jwm{Specifically these experiments focus on} a process known as translocation, the precise motion of the ribosome along its messenger RNA (mRNA) template as it synthesizes the protein encoded by the mRNA (for a review see \cite{tinoco_gd_2012}).

The hypothesized mechanism for translocation (figure \ref{fig_translocation_smfret}a) breaks down into 3 kinetic steps. 
The first step is a thermally driven, reversible transition between two conformational states of the pre-translocation \rlg{ribosomal complex} (labeled GS1 and GS2). 
This transition \rlg{is followed by} the binding of a \rlg{GTPase translation} factor, EF-G, \jwm{which has the effect of stabilizing the GS2 long enough to enable} a GTP hydrolysis-driven movement of the ribosome along the mRNA template. 
The first two steps in this kinetic scheme can be studied experimentally by substitution of GTP with a non-hydrolyzable GTP analogue, preventing the GTP hydrolysis-driven final step in the translocation reaction. 
Figure \ref{fig_translocation_smfret}b shows two time series that measure \rlg{reversible transitions between the GS1 and GS2 states}, customarily plotted as the intensity ratio of the donor and acceptor fluorophores, which is given by \wei{$E_{\rm FRET} = I_{\rm A} / (I_{\rm A} + I_{\rm D})$}. 
The first, recorded in the absence of EF-G, shows a preference for the GS1 state. 
The second time series, taken from an experiment where 500 nM EF-G was added to the solution, shows a shift of the equilibrium towards the GS2 state.

Analysis of 4 experiments performed with EF-G concentrations of 0 nM to 250 nM are shown in figure \ref{fig_translocation_smfret}c. 
We perform VEB inference with 4 states on the ensemble of traces from all experiments to learn a set of shared states for the entire series. 
In each of the experiments we observe two dominant states, corresponding to the GS2 (cyan) and GS1 (magenta) state respectively.
Histograms of observations, on the top row, show a shift in the GS1-GS2 equilibrium towards the GS2 state as the concentration of EF-G is increased.

\jwm{The marginal probability for each state can be expressed in terms of a free energy $G_k$ via the relation $q(z \!=\! k) \sim \exp[-G_k / k_B T]$. 
An energy relative to other states can be defined as}
\[
    \Delta G_k(A)
    =
    \log 
    \frac{\sum_{k \not= l} A_{kl}}
         {\sum_{k \not= l} A_{lk}}
    ,
\]
allowing formulation of a set of posterior distributions
\[
    q(\Delta G_{n,k} \,|\, \hat{\psi})
    =
    \int_{\Delta G_k(A_n) = \Delta G_{n,k}}
    d A_{n,k} \:
    q(A_{n,k} \,|\, \hat{\psi}) 
    .
\]
Figure \ref{fig_translocation_smfret}d shows the posterior ensemble-averaged distributions on $\Delta G_k$. 
The distribution for the GS1 state exhibits a bi-modal signature, indicative of a mixed population of EF-G bound and unbound molecules.
At low EF-G concentrations, the lower mode dominates, whereas the upper mode takes over at high EF-G concentrations, indicating an energetically favorable GS2 state.
This type of bi-modal signature in the posterior would be difficult, if not outright impossible to discern with existing non-hierarchical approaches.

%% file: sec_modelselection.tex
\label{sec:modelselection}

Our analysis of simulated data in section \ref{sec:validation} suggests that the VEB procedure may \rlg{exhibit} a resistance to overfitting similar to what has been observed in other type II maximum likelihood studies.
In particular, we could supply the method with a larger than sufficient maximum number of states at the hyperparameter level and hope to learn a model that leaves superfluous degrees of freedom unpopulated.
To see how well such an approach works in practice we compare analysis of our experimental dataset in the absence of EF-G to a simulated dataset with similar hyperparameters, i.e.~we first perform inference with a single population on the experimental dataset, and then use the inferred hyperparameters to generate our simulated data.

Results of this analysis are shown in figure \ref{fig_K_scaling}.
Inference on experimental data shows a more or less monotonic increase of the lower bound.
This same trend is visible in the lower bound on held-out data obtained from 10-fold cross-validation.
The reason for this becomes apparent when examining histograms of the observations, shown in figure \ref{fig_K_scaling}c.
On a semi-logarithmic scale normally distributed observables should show a parabolic profile.
The experimental histograms at $K=2$ show significant discrepancies with respect to this distribution, in the form of asymmetries and long tails, which are separated into additional states when inference is performed with a higher number of degrees of freedom.

Given that the parameters for each time series are i.i.d., we can attempt to use a heuristic like the ${\rm BIC} = -2 \Lveb + K(K+5)\log N$ for model selection. 
This yields a model with 4 states.
We now use the estimated hyperparameters to sample a simulated dataset from the generative model.
In the absence of artifacts and discrepancies with respect to the underlying model, the monotonic increase of $\Lveb$ is absent, and we in fact observe an all but infinitesimal decrease when overfitting the data.
This decrease is also present, if not necessarily more pronounced, in the lower bound on held out data obtained from cross-validation.
This qualitatively different outcome is also reflected in the effective number of states, shown in figure \ref{fig_K_scaling}b, which increases much more modestly as a function of $K$ as compared to the experimental case.

In short, these results show that the VEB method does in principle have a built-in resistance to overfitting, in the sense that superfluous states are left unpopulated by the algorithm.
Moreover solutions that do overfit the data exhibit a decreased evidence lower bound.
At the same time analysis of actual biophysical data shows that this feature of the method is in practice outweighed by discrepancies between the data and the hypothesized generative model.
Model selection on real biological datasets is therefore limited by our ability to construct a generative model that is not overly sensitive to such discrepancies.

%% file: sec_discussion.tex

We have formulated an approximate parametric empirical Bayes procedure on HMMs based on variational inference, for learning of a single model from a multitude of biological time series.
The estimation of parameter distributions that lies at the core of this method can be thought of as a mechanism for introducing a ``loose'' sharing of parameters between time series.
This type of approach could therefore be useful in any type of learning scenario where it is desirable to extract a single interpretable model from an ensemble of similar yet heterogeneous time series.

The methodology employed here has a number of desirable features, including increased accuracy of posterior inference and a built-in resistance to overfitting. However its main advantage in the context of single-molecule biophysics is that it enables robust estimation on large numbers of time series.
The learned hyperparameters not only provide a single summary representation of an entire experiment, \jwm{but are also useful in validating mechanistic hypotheses encoded in different graphical model variants}.
Comparison of prior and posterior densities allows detailed evaluation of the agreement between observed data and a chosen graphical model, even in the case where one may not wish to blindly employ the evidence lower bound as a model selection criterion.
For example, the bi-modal signature of the posterior in figure \ref{fig_translocation_smfret} suggests that a more detailed analysis could be performed by assuming a mixture of parameter distributions $p(A \,|\, \psi_m)$ conditioned on a new latent state that indicates the presence of an EF-G molecule bound to the ribosomal complex.

In experimental platforms beyond smFRET, several obvious extensions to graphical models related to the HMM could extend the scope of applications of this VEB approach. 
Double Markov chain models can be used to model diffusive motion governed by a latent state, as observed in tethered particle motion experiments \cite{beausang_bpj_2007}. 
Factorial HMMs \cite{ghahramani_ml_1997}, can be used to model systems with orthogonal degrees of freedom such as the internal state of a molecular machine that performs a stepping motion along a substrate, or dynamic binding and unbinding of external factors.

All code used in this publication is available as open source on \url{http://ebfret.github.com}.

\begin{figure}[!t]
    \noindent\includegraphics[width=\columnwidth]{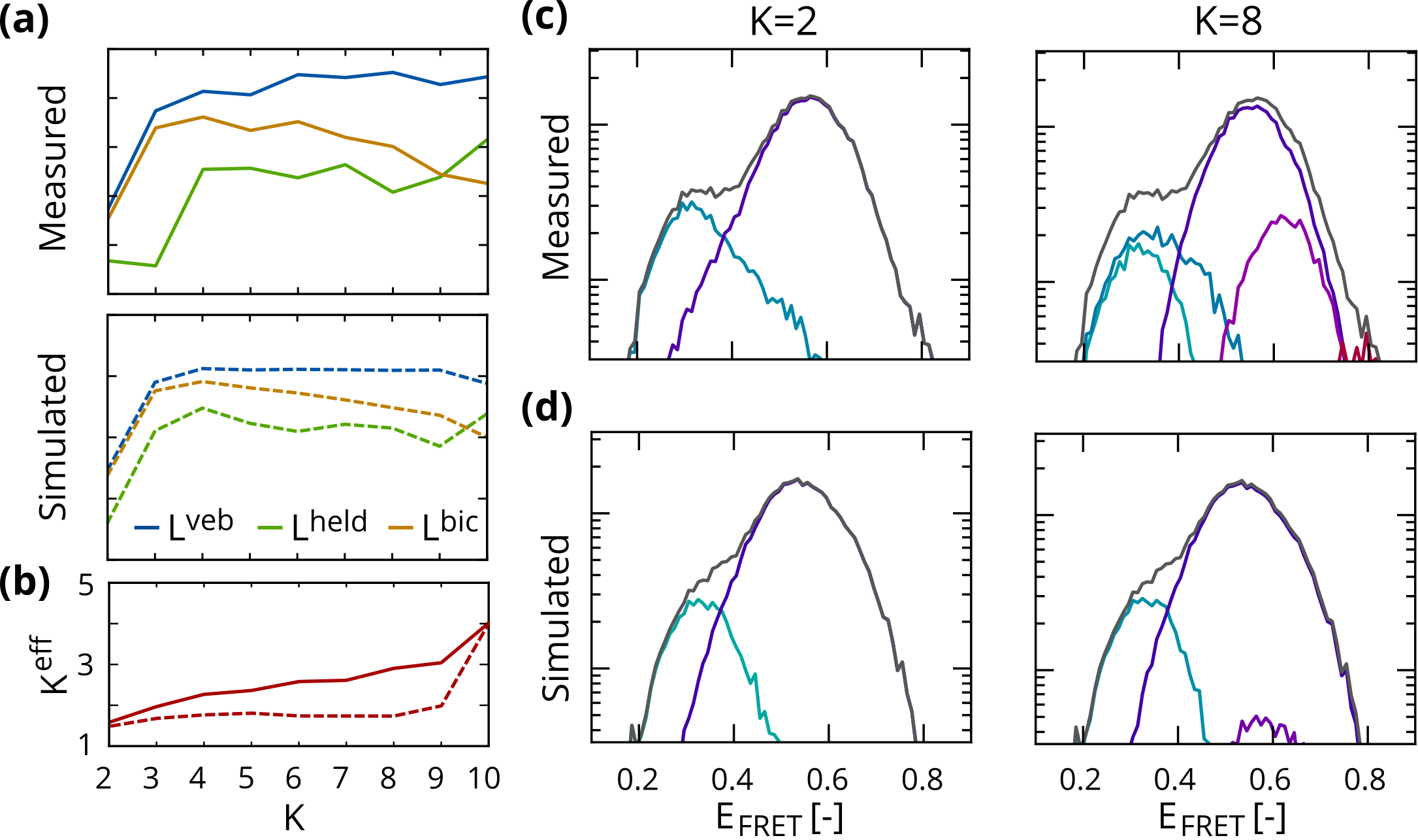}

    \caption{
    \label{fig_K_scaling}
    Dependence on number of states in experimental and simulated data. 
    {\bf (a)} Lower bound $\Lveb$ (blue), lower bound on held out data (green) and heuristic lower bound $L^{\rm bic}$ (orange).
    {\bf (b)} The effective number of states increases monotonically for
    experimental data, but does not reveal significant over-fitting on simulated data.
    {\bf (c)} Experimental data show asymmetries and long tails, resulting
    in population of additional states.
    {\bf (d)} In simulated data, which lack these discrepancies with respect to the generative model, additional states are left largely
    unpopulated.
    }
\end{figure}